\documentclass[runningheads]{llncs}

\usepackage{eccv}
\usepackage{eccvabbrv}
\usepackage{graphicx}
\usepackage{booktabs}
\usepackage{subcaption}
\usepackage[accsupp]{axessibility}  
\usepackage{hyperref}
\usepackage{algorithm}
\usepackage{algpseudocode}
\usepackage{amsmath}
\usepackage{amssymb}
\usepackage{multirow}

\usepackage[export]{adjustbox}
\usepackage{orcidlink}

\begin{document}

\title{DroneFINE: Domain-Aware Parameter-Efficient Fine-Tuning of
Vision-Language Detectors for Drone Images} 

\titlerunning{DroneFINE}

\vspace{-4mm}
\author{
Ke Wu\inst{1,2,5,6}\orcidlink{0009-0009-1327-985X} \and
Yanan Zhang\inst{3}\orcidlink{0000-0003-1592-1067} \and
Yingjie Gao\inst{2}\orcidlink{0009-0003-4037-4701} \and
Wenhao Li\inst{2}\orcidlink{0009-0001-7791-3321} \and
Chenyu Zhou\inst{2}\orcidlink{0009-0009-1732-8487} \and
Xinzhu Ma\inst{2}\orcidlink{0000-0003-0504-0186} \and
Jiaxin Chen% 
\thanks{Corresponding author.}% 
\inst{2,4} \orcidlink{0000-0002-0112-4166} \and
Di Huang\inst{2}\orcidlink{0000-0002-2412-9330}
}
\vspace{-3mm}

\authorrunning{Wu et al.}

\institute{
National College for Excellent Engineers, Beihang University, Beijing, China
\and
School of Computer Science and Engineering, Beihang University, Beijing, China
\and
School of Computer Science and Information Engineering, Hefei University of Technology, Hefei, China
\and
State Key Laboratory of Virtual Reality Technology and Systems, Beihang University, Beijing, China
\and
Innovation Center for Intelligent System Cognition and Decision-Making, Beijing, China
\and
Information Science Academy of China Electronics Technology Group Corporation, Beijing, China\\
\email{
\{20373041,jiaxinchen\}@buaa.edu.cn}\\
}

\maketitle
\vspace{-1mm}
\vspace{-5mm}
\begin{abstract}
Object detection for Unmanned Aerial Vehicles (UAVs) working in open and dynamic environments is a highly challenging task. While Vision-Language Models (VLMs) have offered a powerful solution for universal object detection, adapting them to UAV scenarios remains non-trivial due to a substantial domain gap between VLM pre-training data and aerial imagery. The prevailing Parameter-Efficient Fine-Tuning (PEFT) methods prove ineffective in bridging this gap, as VLMs’ ``natural-scene, foreground-dominant'' visual priors misalign with the ``bird's-eye-view, background-dominant, small-object'' characteristics of UAV data. To address this issue, we propose DroneFINE, a novel PEFT paradigm comprising two domain-aware complementary modules tailored for VLM-based drone image detectors. Specifically, a data-dependent, foreground-aware, and multi-path adaptation mechanism named HyperAdapter is designed, which overcomes the static structural constraints of PEFT. In addition, a background suppression algorithm named SemanticGate is developed, which is a text-conditioned guidance strategy that employs background vocabulary to actively guide the model in suppressing responses from irrelevant regions. Extensive experiments on VisDrone and UAVDT demonstrate that DroneFINE significantly outperforms existing PEFT methods and achieves performance comparable to full fine-tuning while substantially reducing the fine-tuned parameters.
\vspace{-3mm}
\end{abstract}
\vspace{-5mm}    
\section{Introduction}
\label{sec:intro}

\begin{figure}[t] 
  \centering
  \includegraphics[width=0.95\columnwidth]{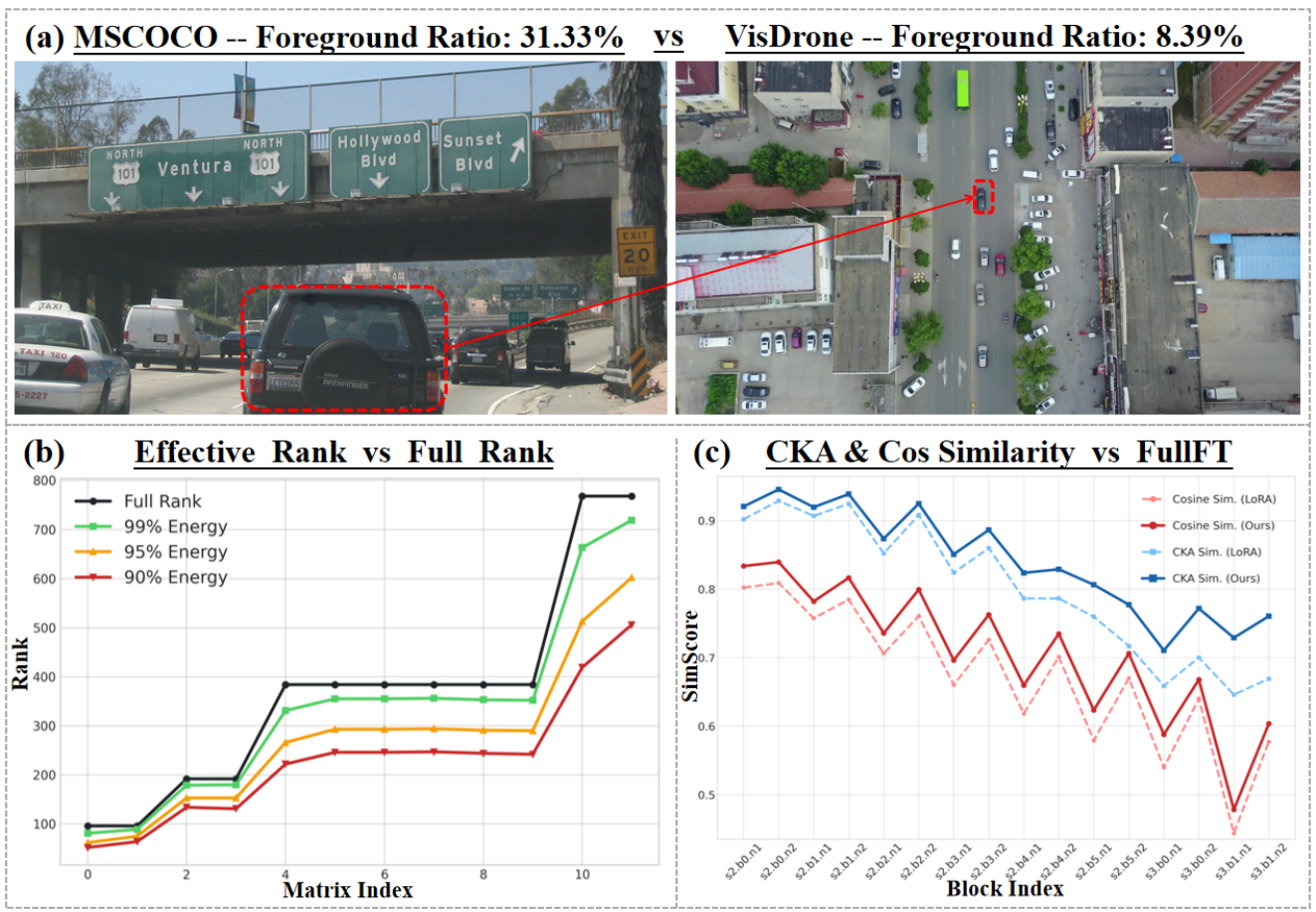} 
    \caption{\textbf{Analysis of Domain Gap and Adaptation Bottlenecks for UAVs.}
      \textbf{(a)} Comparison of image characteristics between common ground-level datasets~\cite{lin2014microsoft} and  aerial datasets~\cite{du2019visdrone}. 
      \textbf{(b)} Rank-Energy($\|A\|_F^2$) plot of FFN layers' weights after Full Fine-Tuning (Full FT). 
      \textbf{(c)} Feature similarity comparison between Full FT, standard LoRA~\cite{hu2022lora}(dashed line), and our method (solid line). 
      \vspace{-4mm}
    }
    \label{fig:vlmfinetune} %
\end{figure} 

UAVs are playing an increasingly vital role in diverse applications such as intelligent inspection, emergency rescue, and logistics delivery, where object detection\cite{cheng2024yoloworld,gao2024ps,gao2026test} serves as a critical perception task to enable these functions~\cite{zhou2025ovodreview,liu2023aerialvln,wang2024traveluav,zhang2025citynavagent,tian2025uavsmeetllms}.
Recently, VLMs~\cite{li2022glip,cheng2024yoloworld,liu2024groundingdino} have shown remarkable generalization and strong adaptability in universal object detection. Therefore, applying the broad knowledge of VLMs to drone detectors is expected to strengthen the robust perceptual capabilities of UAVs operating in complex and dynamic environments.

However, a fundamental domain gap between VLM pre-training data and aerial imagery prevents VLMs from leveraging their inherent strengths when directly applied to UAV scenarios. As shown in Fig.~\ref{fig:vlmfinetune}(a), the visual priors from foreground-dominant data~\cite{shao2019objects365,li2022glip,wang2023v3det,gupta2022grit} generalize poorly to UAV imagery, which typically features bird's-eye-view and small objects in background-dominant scenes. While full fine-tuning or retraining~\cite{pan2025laedino, li2025spar} might boost performance on UAV data, they often cause overfitting, representation collapse, or high computational costs, defeating our original purpose of using VLMs.

In this work, we investigate adapting VLMs to UAV scenarios via popular PEFT methods. Yet, our empirical study reveals that standard PEFT methods yield unsatisfactory transfer results. Furthermore, we analyze and summarize two critical flaws in these conventional methods:
(i) Existing PEFT methods typically update weights in a low-rank space with limited adaptation capability. However, as shown in Fig.~\ref{fig:vlmfinetune}(b), fine-tuning a model for UAV imagery requires a high-rank space due to the complexity of both data and task, particularly in deeper layers. 
(ii) The background-dominant nature of UAV imagery inevitably introduces significant noises from background, making it challenging to adapt the model to pay more attention to visual information in foreground areas.

To address the above issues, we propose a novel PEFT approach dubbed DroneFINE tailored for UAV imagery, by expanding the representation capability while effectively suppressing the interference from background. Specifically, we develop a dynamic adapter HyperAdapter, which strengthens the learning capability by utilizing queries to aggregate foreground features and generating multi-path depthwise separable convolutional kernels. Moreover, we design SemanticGate to unleash the generalization potential of the text branch in VLMs, which utilizes background text features to guide the learning of attention and employs a contrastive learning strategy to discriminate between foreground and background. Extensive experiments on UAV datasets show that DroneFINE outperforms existing PEFT methods by a large margin with minimal overhead, while achieving performance comparable to full fine-tuning.
As shown in Fig.~\ref{fig:vlmfinetune}(c), the feature representations after fine-tuning via our method are significantly more aligned with  full fine-tuning than LoRA~\cite{hu2022lora}. This superior adaptation leads to a 1.7\% mAP and 2.0\% mAP50 improvement over LoRA, 
matching the performance of full fine-tuning on VisDrone~\cite{du2019visdrone}. Furthermore, our method achieves competitive performance with previous SOTA models on UAV imagery, while exhibiting stronger generalization and anti-forgetting capabilities.

The contributions of our work are summarized as below:

\begin{itemize}

    \item We propose DroneFINE, a domain-aware PEFT paradigm for VLM-based drone image detectors that effectively addresses the substantial domain gap. 

    \item We design HyperAdapter, which uses foreground-aware features to generate adaptive kernels, overcoming the representation constraints of PEFT.
    
    \item We develop a SemanticGate algorithm, which leverages background vocabulary to effectively mitigate background noise interference. 
    
    \item Extensive experiments on UAV datasets demonstrate that DroneFINE significantly outperforms existing PEFT methods and achieves performance comparable to full fine-tuning while reducing the fine-tuned parameters. 
    
\end{itemize}

\section{Related Works}
\label{sec:relatedworks}

\subsection{Object Detection on UAV imagery}

Traditional research in UAV imagery object detection, distinct from ground-level, has focused on tackling domain-specific challenges, notably the prevalence of small objects and low foreground-to-background ratios. One line of works attempts to enhance small object discriminability through strategies like multi-scale refinement~\cite{yang2022querydet} or contextual feature enrichment~\cite{du2023ceasc,li2025remdet}. Another line of works focuses on mitigating background noise by explicitly separating foreground regions using techniques such as cluster-based proposals or density-aware sampling~\cite{yang2019clusdet,huang2022ufpmpdet,hu2025domedetr}.
Despite progress on specific datasets, these methods share a fundamental flaw: an insular, closed-set design. They are tightly coupled to their training distribution, not only in terms of predefined object categories but also specific scene characteristics. Consequently, adapting them to new environments or objects requires a cumbersome and resource-intensive modification process. 
This rigidity makes it highly inefficient to tailor UAV models for varying task-specific requirements.
In sharp contrast, our work leverages VLMs as a powerful, feature-rich foundation to overcome this architectural rigidity. We demonstrate that rather than engineering complex, domain-specific modules, the robust representational power of VLMs can be directed towards targeted UAV tasks through remarkably simple adaptation mechanisms. This approach charts a path for an efficient and elegant paradigm: utilizing a strong foundation model alongside a widely applicable adaptation strategy to solve specific UAV detection challenges.
\vspace{-5mm}
\subsection{VLM Adaptation}
VLMs with their strong image-text alignment and generalization capabilities, offer a viable path to address the traditional models' bottlenecks. VLM detectors identify objects via vision-language matching, enabling the detection of a vast range of objects. GLIP~\cite{li2022glip} pioneered VLM detectors through large-scale pre-training by fundamentally unifying object detection and phrase grounding. The GroundingDino~\cite{liu2024groundingdino} series introduced cross-modal attention to DINO~\cite{dino}, becoming a popular base model. YOLO-World~\cite{cheng2024yoloworld} integrated CLIP~\cite{radford2021clip} with YOLOv8~\cite{yolov8}, gaining significant traction due to YOLO's widespread adoption. To apply these powerful, large-scale models to downstream tasks, PEFT methods~\cite{hu2022lora,houlsby2019adapter,zhou2022coop,liu2024shine,liusurrogate,liu2025frequency} have emerged as the key technology. By fine-tuning only a small subset of parameters, PEFT has achieved considerable success in adapting VLMs for general, ground-level vision tasks, effectively mitigating the high overhead and catastrophic forgetting associated with full fine-tuning.

However, adapting VLMs to the unique UAV domain remains an under-explored and challenging area. Several works have attempted to adapt VLMs to the aerial domain via full retraining, such as SPAR~\cite{li2025spar} and LAE-DINO~\cite{pan2025laedino}. Although these methods demonstrate strong performance, their reliance on massive data and computation renders them impractical for rapid iteration. Furthermore, this heavyweight approach is prone to overfitting on limited domain-specific data and risks catastrophic forgetting of the VLM's invaluable pre-trained knowledge. Given the drawbacks of full training, a natural alternative is to apply general PEFT methods directly. However, our preliminary experiments confirm that standard methods like LoRA~\cite{hu2022lora} and Adapter~\cite{houlsby2019adapter} perform poorly on UAV imagery. 
This suggests a critical limitation: the simple representational structures of general PEFT methods are insufficient to bridge the fundamental domain gap between ground-level and aerial data.
Based on this, we propose DroneFINE, a novel PEFT paradigm specifically designed to efficiently and effectively adapt VLMs to the challenging UAV domain by enhancing representational capacity and actively suppressing interference.
\vspace{-3.5mm}

\section{Methodology}
\label{sec:methodology}
\begin{figure*}[t]
  \centering
  \includegraphics[width=0.99\textwidth]{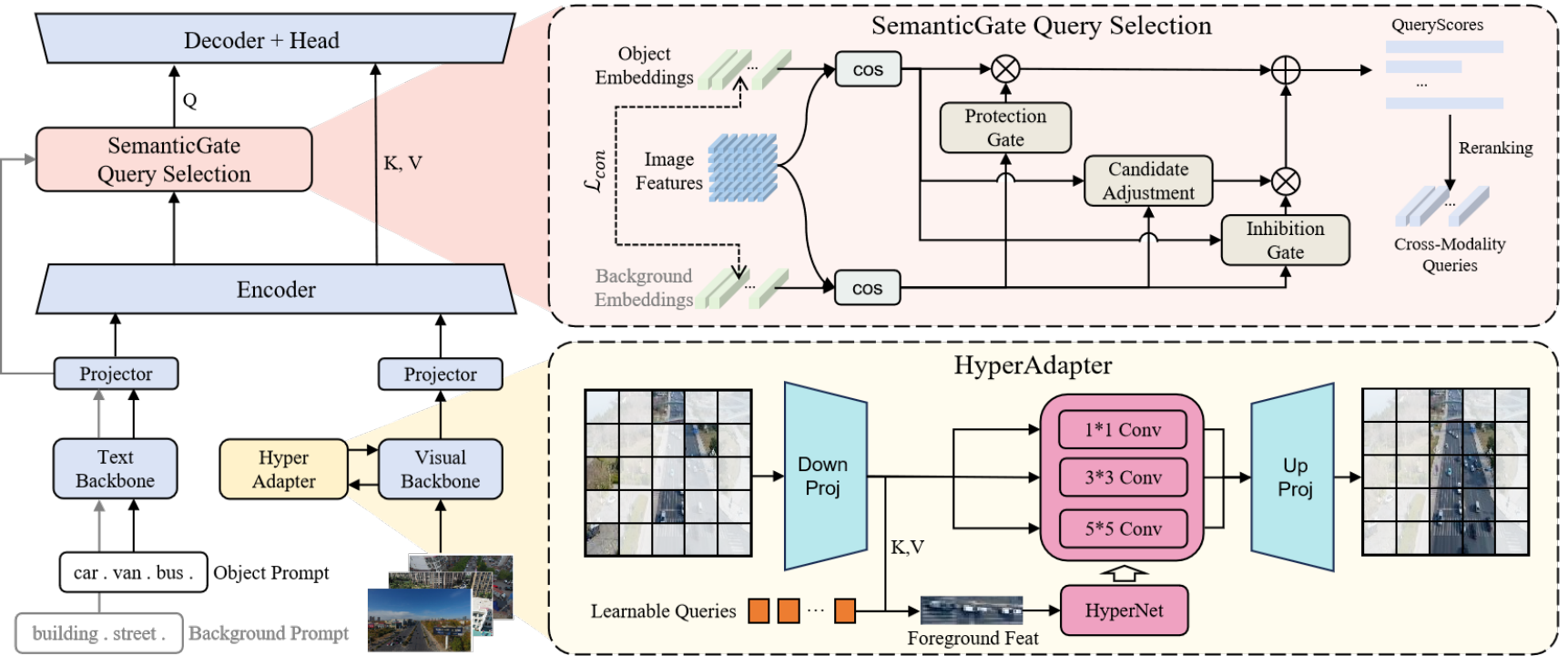}
    \caption{\textbf{Framework of DroneFINE.} The model based on GroundingDINO comprises five parts: (i) frozen visual and text backbones for feature extraction, (ii) projectors for cross-modal alignment, (iii) a detection decoder and head, (iv) \textbf{HyperAdapter} as a serial adapter for the visual backbone, and (v) \textbf{SemanticGate} for enhanced GroundingDINO's language-guided query selection.}
  \label{fig:overview}
\end{figure*}
 
\subsection{Framework Overview}
As illustrated in Fig.~\ref{fig:overview}, our method builds upon the pre-trained vision-language object detector, GroundingDINO~\cite{liu2024groundingdino}.
To improve the model's performance on UAV imagery, we introduce two novel modules dedicated to effective domain adaptation. Specifically, we propose the HyperAdapter to bolster the visual backbone's representational capacity via \textbf{foreground-aware} dynamic multi-path feature extraction. This is complemented by the SemanticGate, which refines the query selection mechanism through suppressing \textbf{background interference} and re-ranking queries, thus focusing the model's attention on the objects.

\subsection{Preliminary} 

Full fine-tuning is the most intuitive approach for adapting a pre-trained vision-language detector to UAV imagery, as it involves updating all of the model's parameters. Formally, given a dataset $D = \{(x_i, y_i)\}_{i=1}^{N}$, the optimization objective is defined as:

\begin{equation} \label{eq:full_finetune}
\theta \leftarrow \arg\min_{\theta} \mathcal{L}(D, \theta),
\end{equation}
where $\theta$ denotes all the parameters, and $\mathcal{L}$ denotes the training loss of the model.

Despite its simplicity and effectiveness, full fine-tuning remains computationally expensive and prone to overfitting. To address these limitations, we adopt the parameter-efficient paradigm of adapter-tuning~\cite{houlsby2019adapter}. Adapter-tuning freezes the pre-trained backbone and updates the lightweight, inserted adapter modules, its optimization objective is:
\begin{equation} \label{eq:adapter_tuning}
\omega \leftarrow \arg\min_{\omega} \mathcal{L}(D, \theta_{F}, \omega),
\end{equation}
where $\theta_{F}$ denotes the fixed backbone parameters, and $\omega$ represents the trainable parameters, including the adapters and the remaining parts of the detector. 

\subsection{HyperAdapter}

\begin{figure}[t]
  \centering
  \includegraphics[width=0.99\columnwidth]{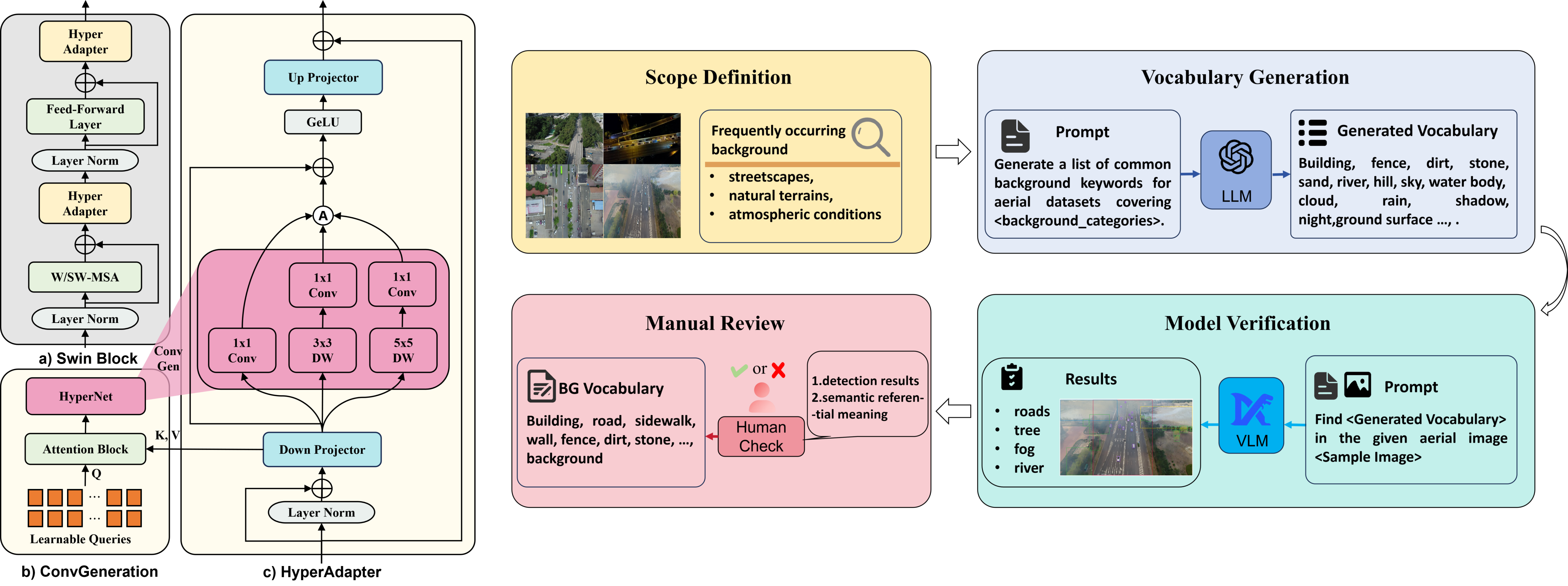} 
    \caption{\textbf{(L)Architecture of HyperAdapter}: Similar to a conventional visual adapter, the HyperAdapter is positioned after the attention and MLP layers. It introduces a foreground-aware multi-path convolution generation mechanism (highlighted in red in the figure) into the adapter. This mechanism utilizes queries to aggregate foreground features, which are then used to generate convolutions. These convolutions, in turn, leverage inductive bias for enhanced feature extraction.\textbf{(R)Process of Background Vocabulary Generation}:A background vocabulary is generated across streetscapes, natural terrains, and atmospheric conditions based on background distribution analysis of sampled images. It is validated via DINO-X~\cite{ren2024dino-x} and manual check to ensure detectability and semantic clarity, resulting in 20 finalized vocabulary items.}
    \label{fig:hyperadapter_arch}
    \label{fig:background_vocabulary}
    \label{fig:method_details}
    \vspace{-6.5mm}
\end{figure} 

\noindent \textbf{Motivation.}
The profound domain gap between VLM pre-training data and UAV imagery is fundamentally characterized by dramatic shifts in object appearance. Beyond this profound domain gap, UAV imagery exhibits high intra-domain variability, where the appearance, scale, and spatial distribution of objects fluctuate markedly across samples. Specifically, under bird's-eye-view, the extremely limited scale of densely distributed targets (\eg, pedestrians, bicycles) reduces them from textured objects into flattened geometric contours, rendering them as mere sparse pixel clusters. Furthermore, dominant environmental backgrounds overwhelm these already fragile features, making them nearly indistinguishable from noise. As observed in our experiments, this combination of fragile visual features and massive intra-domain variability causes VLMs to fail predominantly on small and densely distributed objects.

\noindent \textbf{Conventional Architecture Limitations.}
Effective adaptation methods must possess sufficient capacity to reconstruct robust visual features against this representational collapse. As our Singular Value Decomposition (SVD) analysis illustrated in \cref{fig:vlmfinetune}(b), when decomposing the weight matrices of the FFNs via SVD, a remarkably high rank is required to maintain a low approximation error under UAV imagery. This indicates that simple low-rank linear approximations cannot adequately capture the complex spatial details of aerial objects. While incorporating static convolutional operations~\cite{jie2022convpass,yin2025mona} moderately alleviates this issue by introducing spatial inductive biases, they still exhibit critical representational bottlenecks in the UAV domain. Static structures applying a globally fixed set of filters lack the flexibility to accommodate extreme intra-domain variability and severe background noise. Crucially, as evidenced by our experiments on UAVDT, these static kernels lack the ability to dynamically guide the model toward discriminative regions. Instead of selectively emphasizing object-relevant semantics, the fixed parameter space forces a uniform feature extraction paradigm across the entire image. This lack of dynamic routing dilutes the model's focus, causing it to expend valuable representational capacity on redundant background noise.

\noindent \textbf{HyperAdapter's Architecture.}
To tackle these issues, we introduce HyperAdapter, which is characterized by its foreground-aware architecture and multi-path convolution generation. 
Specifically, as shown in \cref{fig:hyperadapter_arch}, HyperAdapter employs learnable foreground queries to extract essential foreground information. Because these queries are optimized across the entire training set, they encapsulate dataset-level inter-image priors, while their cross-attention with individual feature maps extracts specific intra-image contexts.
Subsequently, these foreground features guide the generation of multi-path convolutions tailored for multi-scale object handling. This dynamic process substantially boosts the representation capacity of traditional low-rank adapters.
As demonstrated by the feature visualization in \cref{fig:foreground_vis}, our mechanism successfully guides the model to focus precisely on the foreground regions, effectively verifying its capability to capture objects amid complex backgrounds.
The complete computation flow of HyperAdapter is defined as:
\begin{align}
\label{eq:hyper_gen}
\mathbf{W}_{dyn} &= H(\mathbf{z}), \\
\label{eq:flat_x}
\mathbf{x}^{\mathrm{flat}} &= \text{Reshape}\left(\mathbf{W}_{down}\tilde{\mathbf{x}}\right),\\
\label{eq:hyper_conv}
\mathbf{p}_{out} &= \text{Reshape} \left(\text{DynConv}\left(\mathbf{x}^{\mathrm{flat}}; \mathbf{W}_{dyn}\right)\right), \\
\label{eq:hyper_out}
\mathbf{x_{out}} &= \mathbf{x} + \mathbf{W}_{up}\left(\text{GELU}\left(\mathbf{p}_{out}\right)\right),
\end{align}
where $\tilde{\mathbf{x}}$ denotes the scaled input features as in~\cite{yin2025mona}, which are subsequently projected into the low-rank space by $\mathbf{W}_{down}$. The hypernetwork $H$~\cite{ha2016hypernetworks} maps the foreground feature $\mathbf{z}$ to data-dependent weights $\mathbf{W}_{dyn}$, which parameterize the $\text{DynConv}$ module comprising parallel $1\times1$, $3\times3$, and $5\times5$ branches. Finally, the refined features are projected back by $\mathbf{W}_{up}$ and added to the input $\mathbf{x}$.

\noindent \textbf{Foreground-Aware Module.}
To improve object perception, we propose leveraging aggregated foreground information to generate feature extractors. This ensures that the generated features can perform thorough extraction on regions that are semantically similar to the foreground cues. Given the background dominance in UAV imagery, we utilize $M$ learnable queries $\mathbf{q}$ to aggregate foreground features via Cross-Attention instead of simple pooling. First, the attention weights $\mathbf{A}$ are computed between the queries and the flattened input features $\mathbf{x}^{\mathrm{flat}}$:
\begin{equation} \label{eq:attn}
\mathbf{A} = \operatorname{softmax}\left( \mathbf{q}(\mathbf{x}^{\mathrm{flat}})^T \right) \in \mathbb{R}^{B \times M \times HW}.
\end{equation}
Then, we perform a weighted aggregation of spatial features based on $\mathbf{A}$ and average the resulting $M$ vectors to obtain the global foreground features $\mathbf{z}$. For the $i$-th sample within the same batch, this process is formulated as:
\begin{equation} \label{eq:z_agg}
\mathbf{z_i} = \frac{1}{M} \sum_{j=1}^{M} \left( \sum_{k=1}^{HW} \mathbf{A}_{i,j,k} \cdot \mathbf{x}^{\mathrm{flat}}_{i,k} \right).
\end{equation}
This query-based pooling captures global foreground semantics implicitly, avoiding auxiliary computational overhead.

\noindent \textbf{Parameter-Efficient Multi-Path Convolution Generation.} 
With the global foreground features $\mathbf{z}$ extracted, the hypernetwork $H$ transforms them into dynamic weights. Dynamic parameter generation often presents challenges in training stability and parameter efficiency, as validated in recent studies~\cite{chen2025vfmadapter,li2025hyperlora_cv,lv2024hyperlora_nlp}. 
Drawing inspiration from existing studies, we first employ a single shared hypernetwork across all adapter layers to generate convolution kernels. This strategy not only economizes parameters but also ensures effective training by aggregating gradient feedback from multiple depth levels.
Building on this robust foundation, we tailor the architecture to the specifics of UAV imagery. We introduce a multi-path design that serves a dual purpose: it enhances the perception of objects across varying scales and enriches gradient feedback to the generator through diverse spatial processing branches. 
To minimize the generated parameter count, we also compress the $H$'s hidden dimensions, utilize depthwise separable convolutions, and employ a group-shared parameter mechanism. By dynamically generating a small number of convolution parameters, our method achieves a balance between performance and efficiency.
\vspace{-2mm}
\subsection{SemanticGate}
\noindent \textbf{Motivation.}
The open-vocabulary capability stands as a distinct advantage of VLMs over traditional models. While standard detection protocols narrow their focus strictly to objects of interest, they often overlook the VLMs' inherent capacity to recognize non-target background contexts. In UAV imagery, we observe a critical asymmetric feature alignment: while foreground features struggle to align with text prompts due to severe domain gaps, dominant background elements remain highly detectable. 

\noindent \textbf{Background Vocabulary Generation.}
Recognizing that the model allocates finite attention resources across the image, we propose a counter-intuitive strategy: instead of passively ignoring the background, we explicitly identify and actively suppress these background responses. This operation explicitly redirects the model's finite attention toward the more challenging foreground objects. As shown in \cref{fig:background_vocabulary}, we establish this systematic background prior through a dedicated four-stage pipeline. To ensure comprehensive semantic coverage, we first sample representative UAV images to analyze common background distributions, identifying three dominant contexts: streetscapes, natural terrains, and atmospheric conditions. Based on these empirical observations, we prompt a LLM(GPT-4) to generate a diverse set of candidate keywords. Since not all textual concepts are visually salient from a bird's-eye view, we subsequently evaluate their empirical detectability by grounding these candidates in real UAV images using DINO-X~\cite{ren2024dino-x}. Finally, a strict manual review is conducted to filter out terms with unstable detection results or ambiguous referential meanings. This rigorous process yields 20 finalized vocabulary items (\eg, "road", "building"), whose embeddings are pre-computed to minimize runtime computational costs. Details of this generation process are provided in the Supplementary Material.

\noindent \textbf{Gating Query Selection.}
Original query selection in GroundingDino is a mechanism similar to anchor box generation. It influences the position of the queries, which effectively dictates the model's regions of attention within the image. To redirect the model's attention to foreground regions, we propose SemanticGate Query Selection. 
We compute the similarity between background text features $\mathbf{F}_{back}$ and image features $\mathbf{F}_{img}$, resulting in a max background similarity score, $\mathbf{s}_{bg}$. This is compared against the max original (foreground) score, $\mathbf{s}_{o}$. As shown in Algo.~\ref{alg:query_selection_suppressed}: \begin{align} \label{eq:s_o_s_bg} \mathbf{s}_o = \max(\mathbf{S}_{orig}) \quad \text{and} \quad \mathbf{s}_{bg} = \max(\mathbf{S}_{back}), \end{align} where $\mathbf{S}_{orig}$ and $\mathbf{S}_{back}$ are the similarity matrices in Algo.~\ref{alg:query_selection_suppressed}. Specifically, $\mathbf{S}_{orig} \in \mathbb{R}^{N_{img} \times N_{text}}$ and $\mathbf{S}_{back} \in \mathbb{R}^{N_{img} \times N_{back}}$, where $N_{img}$, $N_{text}$, and $N_{back}$ denote the number of image, text, and background tokens respectively. 

\begin{algorithm}[t]
\caption{SemanticGate Query Selection: Enhanced GroundingDino's Language-guided Query Selection with Background Suppression}
\label{alg:query_selection_suppressed}
\begin{algorithmic}[1]
    \State \textbf{Input:} $\mathbf{F}_{img}$, $\mathbf{F}_{text}$, $\mathbf{F}_{back}$, $K$
    \State \textbf{Output:} $\textit{topk\_idx}$
    
    \State $\mathbf{S}_{orig} \gets \texttt{torch.einsum("bic,btc->bit", } \mathbf{F}_{img}, \mathbf{F}_{text}\texttt{)}$
    \State $\mathbf{s}_o \gets \mathbf{S}_{orig}\texttt{.max(dim=-1)[0]}$
    
    \State $\mathbf{S}_{back} \gets \texttt{torch.einsum("bic,btc->bit", } \mathbf{F}_{img}, \mathbf{F}_{back}\texttt{)}$
    \State $\mathbf{s}_{bg} \gets \mathbf{S}_{back}\texttt{.max(dim=-1)[0]}$
    
    \State $\mathbf{S}_{adj} \gets \text{SemanticGateModule}(\mathbf{S}_{orig}, \mathbf{s}_o, \mathbf{s}_{bg})$
    \State $\mathbf{s}_{adj} \gets \mathbf{S}_{adj}\texttt{.max(dim=-1)[0]}$
    \State $\textit{topk\_idx} \gets \texttt{torch.topk}(\mathbf{s}_{adj}, K, \texttt{dim=1})[1]$
    \State \textbf{return} $\textit{topk\_idx}$
\end{algorithmic}
\end{algorithm}

To utilize these similarities, we adopt a gating mechanism inspired by traditional LSTM~\cite{hochreiter1997lstm}'s forget, input, and output gates. It is designed to suppress query candidates where the specific background score $s_{bg\_can}$ is high and the foreground score $s_{o\_can}$ is low, indicating a probable background patch.

To resolve the competitive tension between $\mathbf{s}_{bg}$ and $\mathbf{s}_{o}$, we formulate a dynamic gating mechanism comprising three synergistic components: a Protection Gate (PG), an Inhibition Gate (IG), and a Candidate Adjustment (CA) module. PG evaluates the foreground confidence to determine the retention ratio of the original score $\mathbf{S}_{orig}$; for candidates where $s_{o\_can}$ is high, $\mathbf{g}_{pg}$ approaches 1, safely "protecting" potential target queries. Conversely, IG controls the activation of the suppression mechanism; when a candidate exhibits distinct background dominance (high $s_{bg\_can}$ and low $s_{o\_can}$), $\mathbf{g}_{ig}$ approaches 1 to enable suppression. Finally, CA quantifies the magnitude of the potential score adjustment, which is bounded by a $\tanh$ function and scaled by a hyperparameter $\alpha$. Formally, these components are jointly defined as:
\begin{equation} \label{eq:semantic_gates}
\begin{aligned}
\text{(PG)} \quad \mathbf{g}_{pg} &= \sigma(w_{pg}^{bg} \cdot \mathbf{s}_{bg} + w_{pg}^{o} \cdot \mathbf{s}_{o} + b_{pg}), \\
\text{(IG)} \quad \mathbf{g}_{ig} &= \sigma(w_{ig}^{bg} \cdot \mathbf{s}_{bg} + w_{ig}^{o} \cdot \mathbf{s}_{o} + b_{ig}), \\
\text{(CA)} \quad \mathbf{a}_{sig} &= w_{ca}^{bg} \cdot \mathbf{s}_{bg} + w_{ca}^{o} \cdot \mathbf{s}_{o} + b_{ca}, \quad \mathbf{a}_{adj} = \tanh(\mathbf{a}_{sig}) \times \alpha.
\end{aligned}
\end{equation}
These components construct the final adjusted score matrix $\mathbf{S}_{adj}$, where suppression is applied via $\mathbf{g}_{ig}$ and the original score is preserved via $\mathbf{g}_{pg}$:
\begin{equation} \label{eq:adjusted_score_opt}
\mathbf{S}_{adj} = \mathbf{S}_{orig} \odot \mathbf{g}_{pg} + \mathbf{a}_{adj} \odot \mathbf{g}_{ig}.
\end{equation}
The gating mechanism above critically depends on the assumption that $\mathbf{s}_{bg}$ and $\mathbf{s}_{o}$ are meaningful and distinct. To enforce this separability, we introduce a feature contrast solution. We apply an InfoNCE~\cite{oord2018infonce}-based contrastive loss, $\mathcal{L}_{con}$, on the text embedding features $\mathbf{F}_{text}$ and $\mathbf{F}_{back}$ used for vision-language matching.
\begin{equation} \label{eq:contrast_loss_opt}
\begin{aligned}
\mathbf{F}_{comb} &= \text{Concat}([\mathbf{F}_{back}, \mathbf{F}_{text}]), \\
\mathcal{L}_{con} &= \text{TokenContrastiveLoss}(\mathbf{F}_{comb}).
\end{aligned}
\end{equation}
This loss pushes text tokens of different semantics (\eg "road" vs "car") apart in the embedding space, ensuring that our background suppression mechanism operates on high-quality, separable features. Detailed statistical discussion of SemanticGate is relegated to the Supplementary Material.
\vspace{-4mm}

\section{Experiments}
\label{sec:experiments}

\noindent\textbf{Datasets}:
To validate the performance of our model on aerial imagery, we select two datasets for training and validation: VisDrone~\cite{du2019visdrone} and UAVDT~\cite{du2018uavdt}.
Overall, VisDrone challenges models with densely distributed small objects across diverse scenes, while UAVDT poses a high risk of overfitting due to the severe inter-frame similarity in its video sequences. 

\noindent\textbf{Evaluation Metrics}:
We evaluate model performance using the object detection metrics mAP$_{50}$ and mAP (IoU=0.50:0.95) on the corresponding datasets.

\noindent\textbf{Implementation Details}:
We adopt GroundingDino~\cite{liu2024groundingdino}, implemented with MMDetection~\cite{mmdetection,zhao2024mmgroundingdino}, as our base model. For all fine-tuning strategies, we employ a low-rank space with $dim=64$ to ensure a fair comparison. All adaptation methods are implemented by freezing both the text backbone and the visual backbone, while fine-tuning the decoder and detection head, restricting adjustments only to modality-specific modules.
\vspace{-2mm}

\subsection{Quantitative Evaluation}

\begin{table*}[t]
\centering
\caption{Comparison of accuracy and parameter efficiency with
state-of-the-art fine-tuning approaches on VisDrone and UAVDT.}
\label{tab:visdrone_uavdt_comparison}

\vspace{-1.5mm}
\renewcommand{\arraystretch}{0.82}
\setlength{\tabcolsep}{4pt}

\resizebox{0.82\textwidth}{!}{%
\begin{tabular}{l c c c c c c}
\toprule
\multirow{2}{*}{\textbf{Method}} &
\multirow{2}{*}{\textbf{$\Delta$ Params.}} &
\multirow{2}{*}{\textbf{$\Delta$ Trainable.}} &
\multicolumn{2}{c}{\textbf{VisDrone (\%)}} &
\multicolumn{2}{c}{\textbf{UAVDT (\%)}} \\
\cmidrule(lr){4-5}
\cmidrule(lr){6-7}
& & &
\textbf{mAP} &
\textbf{mAP$_{50}$} &
\textbf{mAP} &
\textbf{mAP$_{50}$} \\
\midrule

Zero-shot
& 0 & -- & 16.5 & 26.9 & 9.5 & 18.5 \\

Baseline
& 0 & 0 & 37.1 & 59.6 & 24.7 & 40.9 \\

Full Fine-tuning
& 0 & 27.5M & \textbf{38.4} & \textbf{61.3} & 22.4 & 38.6 \\

\midrule
\multicolumn{7}{l}{\textit{Visual-Side Fine-tuning}} \\

Adapter~\cite{houlsby2019adapter}
& 1.14M & 1.14M & 37.4 & 60.1 & 22.5 & 39.3 \\

LoRA~\cite{hu2022lora}
& 1.13M & 1.13M & 36.7 & 59.3 & \underline{25.0} & 40.8 \\

NormTuning~\cite{giannou2023normtuning}
& 0 & 0.025M & 37.1 & 59.6 & 22.7 & 40.1 \\

Mona~\cite{yin2025mona}
& 1.4M & 1.4M & \underline{37.9} & 60.5 & 22.2 & 37.6 \\

VPT~\cite{jia2022vpt}
& 0.08M & 0.08M & 37.1 & 59.6 & 22.3 & 37.8 \\

\midrule
\multicolumn{7}{l}{\textit{Text-Side Fine-tuning}} \\

CoOp~\cite{zhou2022coop}
& 0.012M & 0.012M & 37.3 & 60.0 & 24.9 & 41.4 \\

Shine~\cite{liu2024shine}
& 0.02M & 0.02M & 37.0 & 60.1 & 24.3 & \underline{42.4} \\

\midrule
\textbf{DroneFINE-T (Ours)}
& 1.54M & 1.54M
& \textbf{38.4}
& \underline{61.2}
& \textbf{26.5}
& \textbf{43.6} \\

\bottomrule
\end{tabular}%
}

\end{table*}

\begin{table*}[t]
\centering
\caption{Performance comparison with UAV-based object detectors on VisDrone (left) and UAVDT (right). Other VLMs are pre-trained on the LAE-1M dataset and fine-tuned on the corresponding dataset, unlike our proposed DroneFINE.}
\label{tab:combined_sota}
\vspace{-2.5mm}

\renewcommand{\arraystretch}{0.88}
\setlength{\tabcolsep}{4pt}

% 左表：VisDrone
\begin{subtable}[t]{0.51\linewidth}
\centering
\resizebox{\linewidth}{!}{
\begin{tabular}{l c c c}
\toprule
\textbf{Method} &
\textbf{Backbone} &
\textbf{mAP (\%)} &
\textbf{mAP$_{50}$ (\%)} \\
\midrule

\multicolumn{4}{l}{\textit{Traditional UAV Models}} \\
ClusDet~\cite{yang2019clusdet} (ICCV'19)
    & ResNet-50 & 26.7 & 50.6 \\
UFPMP-Det~\cite{huang2022ufpmpdet} (AAAI'22)
    & ResNet-50 & 36.6 & 62.4 \\
CEASC~\cite{du2023ceasc} (CVPR'23)
    & ResNet-50 & 28.7 & 50.7 \\
QueryDet~\cite{yang2022querydet} (CVPR'22)
    & ResNet-50 & 28.3 & 48.1 \\
\midrule

\multicolumn{4}{l}{\textit{Recent SOTA UAV Models}} \\
RemDet-X~\cite{li2025remdet} (AAAI'25)
    & YOLOv8X & \underline{40.0} & 61.9 \\
ESOD~\cite{liu2024esod} (TIP'24)
    & YOLOv5 & 37.9 & 62.3 \\
Dome-DETR~\cite{hu2025domedetr} (ACM MM'25)
    & HGNetv2L & 39.0 & 61.1 \\
\midrule

\multicolumn{4}{l}{\textit{VLMs}} \\
YOLO-World~\cite{cheng2024yoloworld} (CVPR'24)
    & YOLOv8L & -- & 55.3 \\
RT-OVAD~\cite{wei2024ovadet} (arXiv'25)
    & ResNet-50 & -- & \underline{64.6} \\
LAE-DINO~\cite{pan2025laedino} (AAAI'25)
    & Swin-T & -- & 56.4 \\
\midrule

\textbf{DroneFINE-T (Ours)}
    & Swin-T & 38.4 & 61.2 \\
\textbf{DroneFINE-B (Ours)}
    & Swin-B & 38.8 & 61.9 \\
\textbf{DroneFINE-L (Ours)}
    & Swin-L & \textbf{42.3} & \textbf{65.8} \\
\bottomrule
\end{tabular}
}
\end{subtable}%
\hfill%
% 右表：UAVDT
\begin{subtable}[t]{0.45\linewidth}
\centering
\resizebox{\linewidth}{!}{
\begin{tabular}{l c c}
\toprule
\textbf{Method} &
\textbf{mAP (\%)} &
\textbf{mAP$_{50}$ (\%)} \\
\midrule

\multicolumn{3}{l}{\textit{Traditional UAV Models}} \\
ClusDet~\cite{yang2019clusdet} (ICCV'19)
    & 13.7 & 26.5 \\
UFPMP-Det~\cite{huang2022ufpmpdet} (AAAI'22)
    & \underline{24.6} & 38.7 \\
CEASC~\cite{du2023ceasc} (CVPR'23)
    & 17.1 & 30.9 \\
\midrule

\multicolumn{3}{l}{\textit{Recent SOTA UAV Models}} \\
RemDet-L~\cite{li2025remdet} (AAAI'25)
    & 20.6 & 34.5 \\
ESOD~\cite{liu2024esod} (TIP'24)
    & 23.6 & \textbf{47.6} \\
\midrule

\multicolumn{3}{l}{\textit{VLMs}} \\
YOLO-World~\cite{cheng2024yoloworld} (CVPR'24)
    & -- & 35.8 \\
RT-OVAD~\cite{wei2024ovadet} (arXiv'25)
    & -- & 40.5 \\
LAE-DINO~\cite{pan2025laedino} (AAAI'25)
    & -- & 36.5 \\
\midrule

\textbf{DroneFINE-T (Ours)}
    & \textbf{26.5} & \underline{43.6} \\
\bottomrule
\end{tabular}
}
\end{subtable}

\vspace{-3mm}
\end{table*}

We conduct a comparative analysis, comparing our method with several representative and recent PEFT methods in Tab.~\ref{tab:visdrone_uavdt_comparison}. Based on these comparisons, we draw the following conclusions:

\noindent\textbf{General VLMs cannot achieve good performance when directly applied to aerial imagery.}
Although VLMs are expected to show strong generalization, the huge domain gap prevents them from achieving good zero-shot performance on UAV images. We observe that categories such as "people", and "awning-tricycle" yield very low mAP$_{50}$ (approx. 2\%-20\%). This highlights the significant discrepancy between aerial images and the VLM's pre-training data.

\noindent\textbf{Text-side adaptation prevents overfitting via semantic alignment but struggles to capture dense, small objects.}
Text-side methods~\cite{zhou2022coop}~\cite{liu2024shine} use minimal parameter updates for high-dimensional semantic alignment. This lightweight approach provides strong regularization, effectively preventing overfitting on highly redundant UAVDT video sequences. However, on VisDrone, such high-level semantic guidance cannot compensate for the lack of low-level visual features required to detect densely distributed small objects. 
To overcome this limitation, DroneFINE jointly optimizes high-level semantics and low-level visual features, achieving stable performance improvements.

\noindent\textbf{Visual-side adaptation is constrained by static representational structures, failing to adapt to the inherent characteristics of extreme data distributions.}
Visual-side methods~\cite{hu2022lora}~\cite{houlsby2019adapter} directly augment spatial features but are restricted by their linear architectures. Even when introducing convs like Mona~\cite{yin2025mona}, the static designs lack the representational flexibility to align with specific data characteristics. Consequently, their representational inadequacy leads to limited performance gains on the spatially complex VisDrone dataset, while triggering severe overfitting on UAVDT, where static kernels fail to adequately capture spatial context.

\noindent\textbf{Our method improves performance by addressing both foreground and background.}
    DroneFINE utilizes a foreground-aware dynamic multi-path convolution and a background suppression module. It achieves 38.4\% mAP and 61.2\% mAP$_{50}$ on VisDrone. For the first time, it matches the performance of the 27.5M-parameter full fine-tuning method by training only 1.54M parameters, which constitutes 5.6\% of the parameters required for full fine-tuning, thereby demonstrating the method's distinct advantage. On UAVDT, DroneFINE achieves gains of  1.8\% in mAP and 2.7\% in mAP$_{50}$ compared to the baseline, which fine-tunes the decoder and detection head while freezing the remaining parameters. Furthermore, it outperforms all existing PEFT methods and even surpasses full fine-tuning on UAVDT.

To more objectively evaluate the effectiveness of our method, 
we conduct performance comparisons with current mainstream detectors for UAV imagery on VisDrone and UAVDT in Tab.~\ref{tab:combined_sota}. On VisDrone, our model DroneFINE-L surpasses the previous SOTA, achieving an improvement of 2.3\% in mAP and 3.4\% in mAP$_{50}$. Compared to other VLM detectors trained on LAE-1M~\cite{pan2025laedino}, our model shows an improvement of approximately 1–10\% in mAP$_{50}$. On UAVDT, our model DroneFINE-T achieves SOTA mAP, surpassing the recent SOTA ESOD~\cite{liu2024esod} by 2.9\%. Although our mAP$_{50}$ is slightly lower, this is because ESOD re-crops the foreground regions, prioritizing foreground detection over precise localization at higher IoU thresholds. Moreover, despite lacking large-scale aerial pre-training, DroneFine matches RT-OVAD~\cite{wei2024ovadet} and LAE-DINO~\cite{pan2025laedino}, confirming the viability of adaptation approaches. Further details regarding the comparison are provided in the Supplementary Material.
\vspace{-2mm}

\subsection{Qualitative Evaluation}

\begin{figure}[t]
  \centering
  \includegraphics[width=0.99\columnwidth]{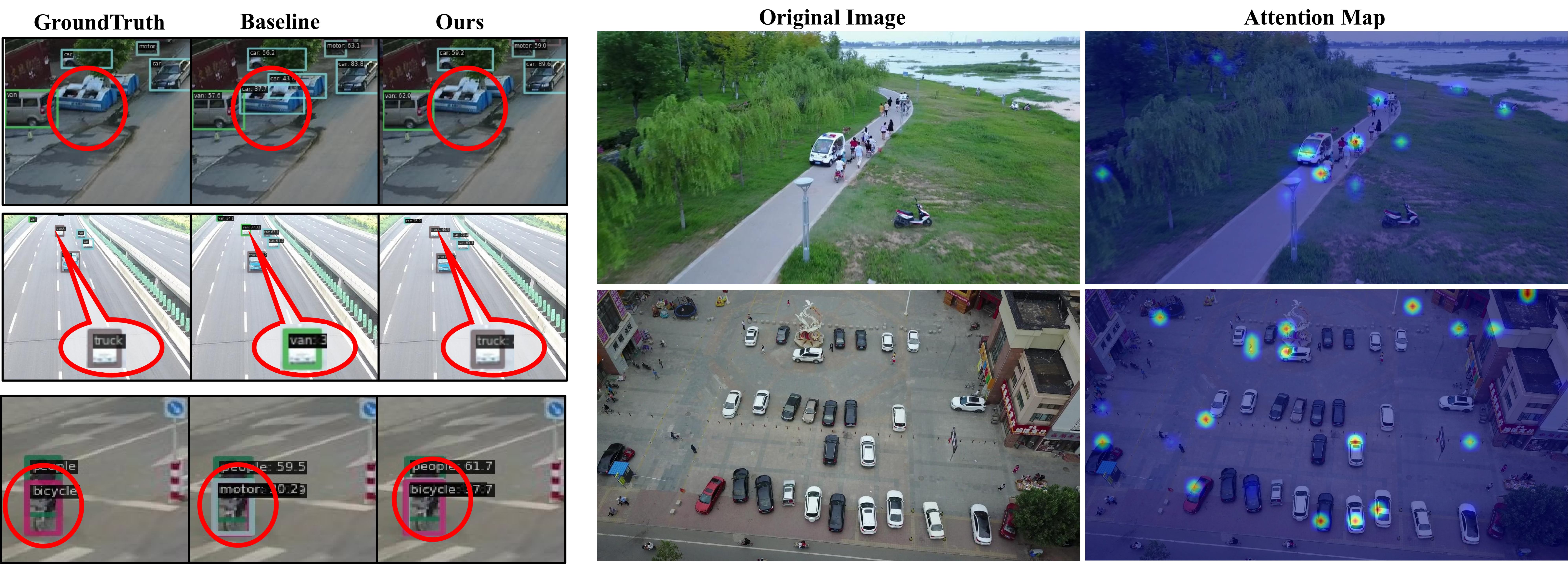} 
    \caption{\textbf{Detection Results and Attention Map on VisDrone.} 
    \textbf{(Left)} Visualization of detection results on the VisDrone. 
    \textbf{(Right)} Visualization of foreground-aware attention maps. Original images and corresponding attention maps generated by HyperAdapter's foreground-aware module.}
    \label{fig:ksh} %
    \label{fig:detection_result}
    \label{fig:foreground_vis}
    \label{fig:foreground_analysis}
\end{figure} 
Fig.~\ref{fig:detection_result} illustrates a qualitative comparison between our method and the baseline on the VisDrone dataset. 
In the first row, the baseline misclassifies dumpsters as cars, incorrectly associating their aerial appearance with the pre-learned "car" concept, highlighting its poor cross-domain alignment. Furthermore, the second and third rows demonstrate the superior robustness of our model on challenging samples. 
Specifically, our method accurately detects the distant truck in the second row and the bicycle in the third row, effectively handling both small objects and rare categories where the baseline fails. In Fig.~\ref{fig:foreground_vis}, warmer colors indicate stronger attention, demonstrating that our foreground-aware mechanism effectively focuses on salient foreground objects such as vehicles and pedestrians.
\vspace{-5mm}
\subsection{Ablation Study}

\begin{table}[t]
\centering
\caption{Ablation study of different components on the VisDrone and UAVDT datasets. ``Full'' indicates the module with all internal mechanisms.}
\label{tab:ablation_main}

\vspace{-1.5mm}
\renewcommand{\arraystretch}{0.82}
\setlength{\tabcolsep}{3pt}

\resizebox{0.7\linewidth}{!}{%
\begin{tabular}{l c c c c}
\toprule
\multirow{2}{*}{\textbf{Method}} &
\multicolumn{2}{c}{\textbf{VisDrone (\%)}} &
\multicolumn{2}{c}{\textbf{UAVDT (\%)}} \\
\cmidrule(lr){2-3}
\cmidrule(lr){4-5}
&
\textbf{mAP} &
\textbf{mAP$_{50}$} &
\textbf{mAP} &
\textbf{mAP$_{50}$} \\
\midrule

Baseline
& 37.1 & 59.6 & 24.7 & 40.9 \\

\midrule
HyperAdapter (Full)
& 38.1 & 61.1 & 26.1 & 41.9 \\

\quad \textit{w/o Foreground Awareness}
& 37.8 & 60.2 & 24.9 & 41.0 \\

\midrule
SemanticGate (Full)
& 37.9 & 60.8 & 26.0 & 42.0 \\

\quad \textit{w/o Contrastive Loss}
& 37.5 & 60.4 & 25.2 & 42.1 \\

\midrule
Mona + CoOp
& 37.5 & 60.3 & 24.5 & 41.7 \\

\textbf{DroneFINE-T (Full Model)}
& \textbf{38.4}
& \textbf{61.2}
& \textbf{26.5}
& \textbf{43.6} \\

\bottomrule
\end{tabular}%
}

\vspace{-3mm}
\end{table}
 
\noindent \textbf{Ablation on the main components.} To analyze the contribution of each component, we conduct ablation studies on the modules and their internal mechanisms across the VisDrone and UAVDT datasets. As shown in Tab. \ref{tab:ablation_main}, the proposed components generally improve the model's fine-tuning performance. Compared to the Baseline, each module consistently increases mAP$_{50}$ by 1.2-1.5\% on VisDrone and 1.0-1.1\% on UAVDT. 
Building on this, we delve deeper into the internal mechanisms of the two modules. The experiments show that dynamically generating convolutions improves performance, but its effect is limited without foreground awareness. The foreground-aware module is crucial, providing a 0.9\% mAP$_{50}$ improvement on VisDrone and UAVDT. Additionally, although removing the contrastive loss reduces overall mAP and mAP$_{50}$ on VisDrone, the standalone SemanticGate still outperforms the baseline. 
Besides, we experiment with a simple combination of text-side (CoOp) and visual-side (Mona) fine-tuning. However, the performance is significantly inferior to DroneFINE, highlighting the complementarity and effectiveness of our method.

\noindent \textbf{Ablation on the Foreground-Aware Module in HyperAdapter.} We also introduce another foreground-aware module, ObjSeeker from ESOD~\cite{liu2024esod}, into the Adapter, which identifies the foreground by predicting a density map. While this module (Adapter + ObjSeeker) slightly improve performance over the baseline, its overall performance is weaker than our proposed HyperAdapter, as shown in~\cref{tab:sub_ablation_auxiliary}. We also attempt to replace the attention scores of HyperAdapter with density map prediction scores. However, this yield no performance improvement. This indicates that effective global aggregated features for highlighting foreground parts can be learned implicitly. Moreover, adding an auxiliary task head to each Adapter significantly increases computational cost. In contrast, our model achieves a better balance between performance and computational overhead. 
Finally, we investigate the impact of the rank on model performance. As shown in~\cref{tab:sub_rank_ablation}, we first conduct an ablation study to identify the optimal rank configuration for HyperAdapter. Furthermore, we compare our approach with other methods across varying rank scales (\cref{tab:sub_rank_scale}), which further demonstrates the effectiveness and parameter efficiency of our proposed method.

\begin{table}[t]
\centering
\caption{Module ablation on VisDrone and additional performance evaluations.
In (d), ``Full'' contains street, terrain, and atmosphere vocabularies,
while ``Random'' denotes target-irrelevant words.}
\label{tab:comprehensive_eval_compact}

\vspace{-3mm}
\renewcommand{\arraystretch}{0.82}
\setlength{\tabcolsep}{2.5pt}

\makebox[\linewidth][c]{%
\begin{subtable}[t]{0.60\linewidth}
\centering
\caption{Foreground-aware module ablation.}
\label{tab:sub_ablation_auxiliary}
\vspace{-2mm}
\scalebox{0.82}{%
\begin{tabular}{@{}l c c@{}}
\toprule
\textbf{Method} & \textbf{mAP} & \textbf{mAP$_{50}$} \\
\midrule
Baseline                 & 37.1          & 59.6 \\
Adapter + ObjSeeker      & 37.3          & 60.0 \\
HyperAda.                & \textbf{38.1} & \textbf{61.1} \\
HyperAda. + ObjSeeker    & 38.0          & \textbf{61.1} \\
\bottomrule
\end{tabular}%
}
\end{subtable}%
\hspace{0.025\linewidth}%
\begin{subtable}[t]{0.32\linewidth}
\centering

\vspace*{1.4mm}

\caption{Rank ablation.}
\label{tab:sub_rank_ablation}
\vspace{-2mm}
\scalebox{0.82}{%
\begin{tabular}{@{}c c c@{}}
\toprule
\textbf{Rank} & \textbf{mAP} & \textbf{mAP$_{50}$} \\
\midrule
32 & 37.3          & 60.0 \\
64 & \textbf{38.4} & \textbf{61.2} \\
96 & 37.9          & 60.9 \\
\bottomrule
\end{tabular}%
}
\end{subtable}%
}

\vspace{0.6mm}

\makebox[\linewidth][c]{%
\begin{subtable}[t]{0.60\linewidth}
\centering
\caption{Rank-scaled methods.}
\label{tab:sub_rank_scale}
\vspace{-2mm}
\scalebox{0.82}{%
\begin{tabular}{@{}l c c c@{}}
\toprule
\textbf{Method} &
\textbf{$\Delta$ Params.} &
\textbf{mAP} &
\textbf{mAP$_{50}$} \\
\midrule
Baseline
    & 0 & 37.1 & 59.6 \\
LoRA ($r=128$)
    & 2.26M & 37.1 & 59.6 \\
Mona ($r=128$)
    & 2.97M & 37.9 & 60.6 \\
\textbf{HyperAda. (Ours)}
    & \textbf{1.54M}
    & \textbf{38.1}
    & \textbf{61.1} \\
\bottomrule
\end{tabular}%
}
\end{subtable}%
}

\vspace{0.6mm}

\makebox[\linewidth][c]{%
\begin{subtable}[t]{0.60\linewidth}
\centering
\caption{Background vocabulary ablation.}
\label{tab:bg_vocab_ablation}
\vspace{-2mm}
{\scriptsize
\setlength{\tabcolsep}{2.2pt}
\renewcommand{\arraystretch}{0.86}

\begin{tabular}{@{}l c c c@{}}
\toprule
\textbf{Vocabulary} &
\textbf{\#Words} &
\textbf{mAP} &
\textbf{mAP$_{50}$} \\
\midrule
Full vocabulary
    & 20 & \textbf{37.9} & \textbf{60.8} \\
Streetscapes
    & 5  & 37.6 & 60.2 \\
Natural terrains
    & 8  & 37.8 & 60.3 \\
Atmospheric conditions
    & 7  & 37.2 & 59.6 \\
\midrule
Random-5
    & 5  & 37.0 & 59.6 \\
Random-10
    & 10 & 36.6 & 59.2 \\
Random-20
    & 20 & 36.9 & 59.4 \\
\bottomrule
\end{tabular}%
}
\end{subtable}%
\hspace{0.025\linewidth}%
\begin{subtable}[t]{0.32\linewidth}
\centering

\vspace*{4.0mm}

\caption{Anti-forgetting \& speed.}
\label{tab:sub_anti_forgetting_fps}
\vspace{-2mm}
\scalebox{0.82}{%
\begin{tabular}{@{}l c c@{}}
\toprule
\textbf{Method} & \textbf{mAP} & \textbf{FPS} \\
\midrule
Full FT       & 42.5          & \textbf{3.22} \\
Mona          & 41.9          & 2.99 \\
\textbf{Ours} & \textbf{44.9} & 2.79 \\
\bottomrule
\end{tabular}%
}
\end{subtable}%
}

\vspace{-3mm}
\end{table}

\noindent \textbf{Analysis of Background Semantics and Query Re-ranking in SemanticGate.} \cref{tab:bg_vocab_ablation} shows that SemanticGate relies on meaningful background semantics rather than arbitrary text tokens. 
Quantitatively, the full vocabulary achieves 37.9\%/60.8\% mAP/mAP$_{50}$, outperforming the best single-category vocabulary by 0.1\%/0.5\% and the size-matched Random-20 vocabulary by 1.0\%/1.4\% points. Moreover, increasing the random vocabulary from 5 to 20 words does not yield consistent gains, indicating that the improvement arises from complementary background semantics rather than vocabulary size.
We also quantify the changes in queries before and after SemanticGate.
On VisDrone, for query selection's top-20 queries, GT-box IoU coverage improves/degrades from pre- to post-SemanticGate on $19.3\%$/$6.0\%$ of images; Recall@0.5 improves/degrades on $6.8\%$/$2.0\%$.
Notably, foreground similarity slightly decreases on $23.9\%$ of images after gating, indicating that the gains are not from simply boosting foreground scores. Instead, SemanticGate uses background scores to re-rank queries and improve GT coverage. A single-gate variant drops $0.7\%$ mAP$_{50}$ on VisDrone, supporting the multi-gate design.

\vspace{-3mm}
\subsection{Efficiency and Anti-Forgetting Evaluation}
To evaluate the practical utility of DroneFINE, we assess its inference efficiency and anti-forgetting capability on COCO. As shown in~\cref{tab:sub_anti_forgetting_fps}, our method achieves a COCO mAP of 44.9\%, significantly outperforming Full FT (42.5\%) and Mona (41.9\%). This demonstrates that DroneFINE effectively preserves the inherent generalization of the VLM.
Compared to the original model, inference speed decreases by only 0.43 FPS on a single 2080 Ti. Considering DroneFINE’s superior performance, this trade-off is entirely reasonable.

\vspace{-3mm}
\section{Conclusion}
\label{sec:conclusion}
Existing PEFT methods fail to bridge the substantial domain gap between ground-level and aerial imagery, limiting the deployment of VLMs in UAV-based object detection. In this paper, we analyze the underlying causes of this failure, attributing it to two main factors: the insufficient representation capacity of original PEFT structures, and the severe interference from complex background noise. To tackle these issues, we propose a novel domain-aware PEFT framework dubbed DroneFINE. Architecturally, DroneFINE employs a foreground-aware dynamic multi-path convolution to significantly enhance foreground representations. Mechanistically, it introduces a background vocabulary coupled with a gating mechanism to effectively suppress background interference. Experimental results demonstrate that DroneFINE significantly improves performance, paving a new path for the application of VLMs in the UAV domain.

\section*{Acknowledgement}
This work was supported in part by the Beijing Natural Science Foundation under Grant 4242044, the National Key R\&D Program of China under Grant 2021YFA1000401, the National Natural Science Foundation of China under Grants U23B2012, U22B2049, and 62506111, the China Postdoctoral Science Foundation under Grant 2025M781468, the Postdoctoral Fellowship Program of the CPSF under Grant GZC20251098, the Research Program of the State Key Laboratory of Virtual Reality Technology and Systems, and the Fundamental Research Funds for the Central Universities.

\bibliographystyle{splncs04}
\bibliography{main}
\end{document}